\pdfoutput=1

\documentclass[11pt]{article}

\usepackage[]{EMNLP2022}

\usepackage{times}
\usepackage{color}
\usepackage{latexsym}
\usepackage{graphicx}
\usepackage{graphics}
\usepackage{subfigure}
\usepackage{multirow}
\usepackage{amsmath}
\usepackage{amssymb}
\usepackage{booktabs}
\usepackage{array}

\usepackage[T1]{fontenc}

\usepackage[utf8]{inputenc}

\usepackage{microtype}

\usepackage{inconsolata}
\usepackage{breakurl}
\usepackage{svg}
\usepackage{booktabs}
\usepackage{microtype}
\newcommand{\comment}[1]{}
\usepackage{svg}
\usepackage{hyperref}

\newcommand{\Chengyuan}[1]{}

\title{Investigating the Robustness of Natural Language Generation \\
from Logical Forms via Counterfactual Samples}

\author{Chengyuan Liu$^{1}$\footnotemark[1]\hspace{1.5mm}, Leilei Gan$^{2}$\footnotemark[1]\hspace{1.5mm}\footnotemark[2]\hspace{1.5mm}, Kun Kuang$^2$\footnotemark[2]\hspace{1.5mm}, Fei Wu$^{2,3,4}$\\
$^1$School of Software Technology, Zhejiang University \\
$^2$College of Computer Science and Technology, Zhejiang University \\
$^3$Shanghai Institute for Advanced Study of Zhejiang University \\ $^4$Shanghai AI Laboratory \\
\{liucy1, leileigan, kunkuang, wufei\}@zju.edu.cn \\
}
\begin{document}
\maketitle


\renewcommand{\thefootnote}{\fnsymbol{footnote}}
\footnotetext[1]{Equal contribution.}
\footnotetext[2]{Corresponding author.}

\renewcommand*{\thefootnote}{\arabic{footnote}}

\begin{abstract}

The aim of Logic2Text is to generate controllable and faithful texts conditioned on tables and logical forms, which not only requires a deep understanding of the tables and logical forms, but also warrants symbolic reasoning over the tables.
State-of-the-art methods based on pre-trained models have achieved remarkable performance on the standard test dataset. However, we question whether these methods really learn how to perform logical reasoning, rather than just relying on the spurious correlations between the headers of the tables and operators of the logical form. 
To verify this hypothesis, we manually construct a set of counterfactual samples, which modify the original logical forms to generate counterfactual logical forms with rarely co-occurred table headers and logical operators.
SOTA methods give much worse results on these counterfactual samples compared with the results on the original test dataset, which verifies our hypothesis.
To deal with this problem, we firstly analyze this bias from a causal perspective, based on which we propose two approaches to reduce the model's reliance on the shortcut. The first one incorporates the hierarchical structure of the logical forms into the model. The second one exploits automatically generated counterfactual data for training.
Automatic and manual experimental results on the original test dataset and the counterfactual dataset show that our method is effective to alleviate the spurious correlation. Our work points out the weakness of previous methods and takes a further step toward developing Logic2Text models with real logical reasoning ability.

\end{abstract}

\section{Introduction}

\begin{figure}[h!]
    \centering
    \includegraphics[width=\linewidth]{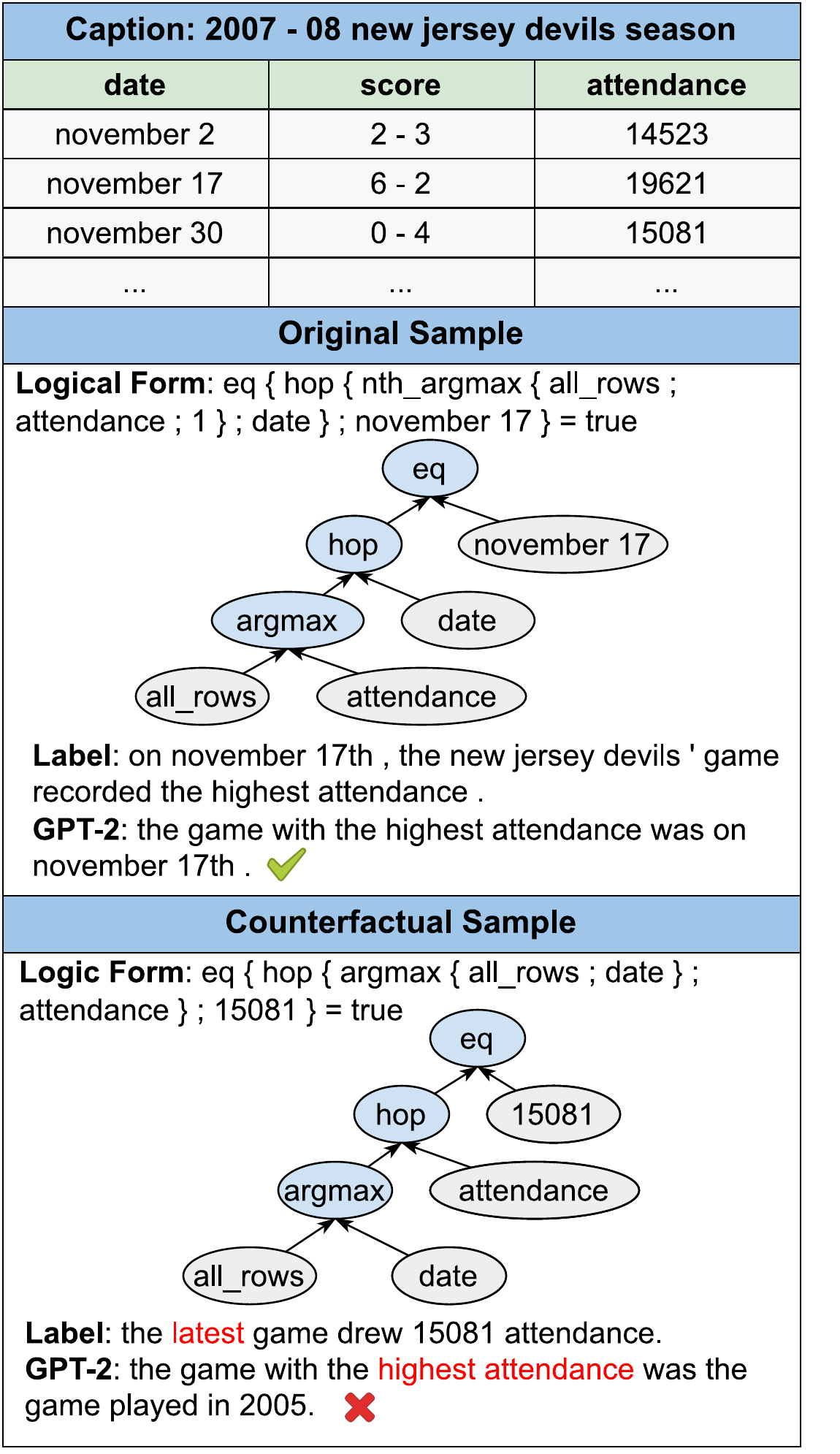}
    \caption{The GPT-2 based model makes a fluent and logical consistent prediction on the original Logic2Text sample, but generates a sentence logically inconsistent with the logical form of the counterfactual sample.}
    \label{fig:sample logical form}
\end{figure}

Recently, generating logical consistent natural language from tables has attracted the attention of the research community \cite{wiseman-etal-2018-learning, DBLP:journals/corr/abs-1806-01353, liang-etal-2009-learning,shu-etal-2021-logic,DBLP:conf/emnlp/ChenSYW20, chen2020logical}.
Given a table, the Logic2Text~\cite{chen2020logic2text} task is required to generate controllable and faithful texts conditioned on a logical form, which not only requires a deep understanding of the tables and logical forms, but also warrants symbolic reasoning over the tables.
State-of-the-art methods based on pre-trained models~\cite{radford2019language, shu-etal-2021-logic} have achieved remarkable performance on the standard test dataset of Logic2Text.
Figure \ref{fig:sample logical form} shows an original sample of the task.

However, we question whether these methods really learn how to perform logical reasoning, rather than just relying on the spurious correlations~\footnote{\textit{Spurious Correlations} or \textit{Shortcuts} refer to connections between two variables that are non-causal in statistics \cite{simon1954spurious}.} between table headers such as ``attendance" and logical operators such as ``argmax".
Several previous studies have demonstrated that such shortcuts severely damage the robustness of models~\cite{branco-etal-2021-shortcutted,wang2021robustness}.

To verify our hypothesis, we manually construct a set of counterfactual samples from the development set and test set of Logic2Text, named LCD (Logical Counterfactual Data)~\footnote{\textit{Counterfactual Samples} are samples that change some variables of the factual samples with the others unchanged \cite{2016Causal}.}. We modify the original logical forms to generate counterfactual logical forms with rarely co-occurred table headers and logical operators, then we annotate the corresponding counterfactual label sentences. Figure \ref{fig:sample logical form} compares the original sample of Logic2Text and the corresponding counterfactual sample. GPT-2 based model makes a fluent and logical consistent prediction on the original sample. However, for the counterfactual sample, the model still employs the “argmax” logical operator to describe “attendance”, which is inconsistent with the logical form. The reason we suppose is that in the training dataset there are a large number of logical forms containing ``$\text{ argmax \{ all}\_\text{rows ; attendance \}}$'', which is used to describe the phrase ``the highest attendance''. A model trained on this biased dataset would learn to exploit such spurious correlations to make predictions, thus failing to perform correct logical reasoning on the counterfactual samples.
We evaluate the state-of-the-art methods on LCD,
and they give much worse results compared with the results on the original test set.

In addition to the bias in the training dataset, previous works directly using linearized logical forms as inputs fail to understand the hierarchical structure of the logical forms, which further aggravates the models to learn the spurious associations between the logical operators and the table headers. 
To deal with this problem, we firstly leverage \textit{Causal Inference}~\cite{2016Causal} to analyze this bias. Based on the analysis, two approaches are proposed: 1) In order to overcome the limitation of linearized logical form inputs, we use different attention masks for different tokens in the logical forms to constrain each token to only interact with the tokens it should reason with; 2) To reduce the reliance on spurious correlations in the training dataset, we train the model on automatically generated counterfactual data, forcing the model to learn real logical reasoning.

Automatic and manual experimental results on the standard test dataset of Logic2Text and LCD demonstrate that our method is able to alleviate spurious correlations and improve logical consistency. Compared with the state-of-the-art baselines, there are 22\% and 14\% less relative decreases after applying our method to GPT-2 and T5, respectively. Our work mainly points out the weakness of current methods, which is easy to be ignored but important for a robust and faithful generation. It takes a further step toward developing Logic2Text models with real logical reasoning ability.

Our codes and data are publicly available at \url{https://github.com/liuchengyuan123/L2TviaCounterfactualSamples}.

\section{Pilot Study on the Robustness of Logic2Text Models}
\label{sec: pilot study}
\subsection{Counterfactual Samples Construction}
To quantify to what extent the bias affects the robustness of Logic2Text models, we manually construct a set of counterfactual samples from the development set and test set of Logic2Text, named LCD (Logical Counterfactual Data). 

\begin{figure}[t]
    \centering
    \includegraphics[width=\linewidth]{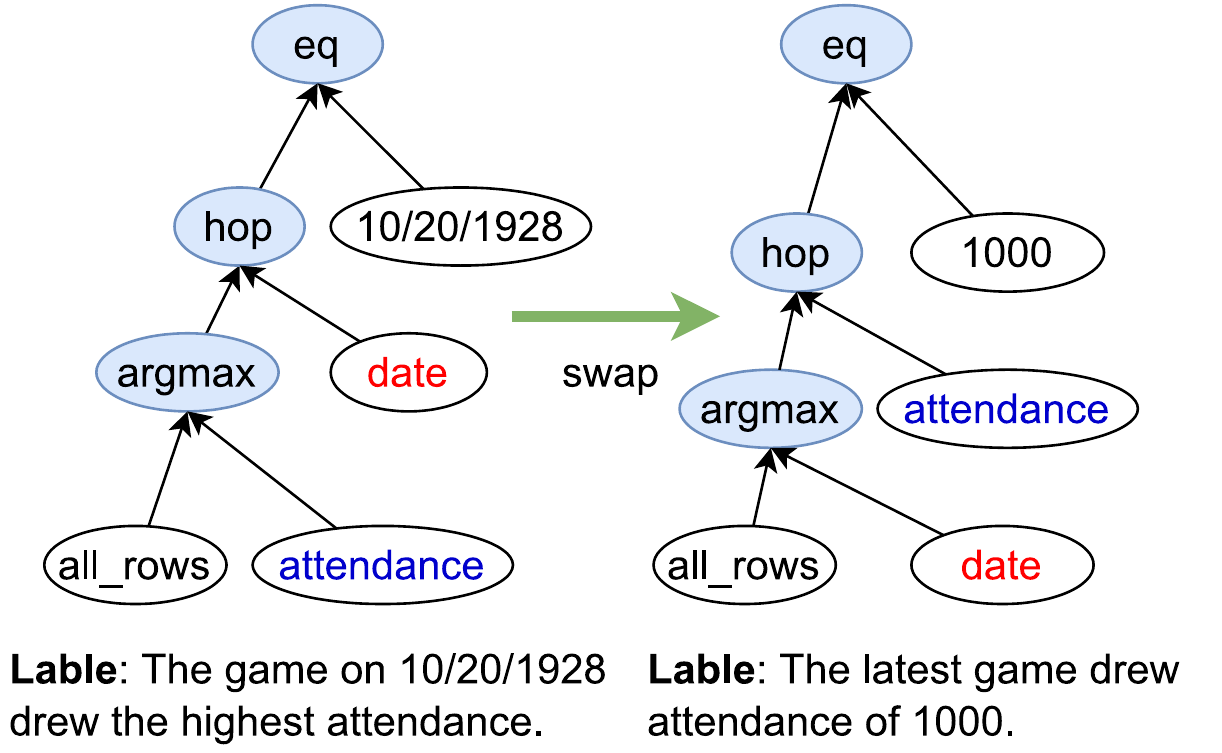}
    \caption{Construction of counterfactual samples.
    Left: Original sample. Right: Counterfactual sample. 
    We construct the counterfactual logical form by replacing the header ``attendance'' with the counterfactual header ``date'', leaving a negative shortcut for the model to exploit. Then we annotate the label sentence based on the counterfactual logical form. 
    \label{fig: LCD_Gen}}
\end{figure}

Specifically, we take datapoints from the development set and test set of Logic2Text and modify the original logical forms to generate counterfactual logical forms with rarely co-occurred table headers and logical operators, then we annotate the corresponding counterfactual label sentences. The tables are left unchanged when constructing counterfactual samples. Figure~\ref{fig: LCD_Gen} shows how to construct a counterfactual sample from the original sample. The ``argmax'' logical operator in the original left sample is applied to the ``attendance'' table header to locate the row in the table. We replace the table header ``attendance'' with a counterfactual table header ``date'' to produce the counterfactual logical form. Then, based on the constructed counterfactual logical form, we annotate the corresponding label sentence.
The reason we choose the ``date'' table header is that after linearizing, both the original logical form and the counterfactual logical form contain the logical operator ``argmax'' and the table header ``attendance'', which leaves a negative shortcut for the models to exploit.


We totally construct 809 counterfactual samples, on which current SOTA Logic2Text models are evaluated.

\subsection{Models}

We evaluate the GPT-2 \cite{radford2019language} and T5 \cite{T5} based Logic2Text models on the counterfactual dataset. These two models achieve the SOTA results on the standard test dataset of Logic2Text.

Formally, given a linearized logical form $L$, and the caption of a table $T$, the input of the pre-trained models is denoted as $X = [P;T;L]$, which is a concatenation of $L$, $T$ and a prefix prompt $P$. The prefix prompt $P$ is ``Describe the logical form: ''. Given a set of training samples $S = \{(L_i, T_i, Y_i)\}|_{i=1}^{N}$ where $Y$ is the label sentence and $N$ is the number of samples, GPT-2 or T5 based Logic2Text models are trained by maximizing the following objective function:
\begin{equation}
    J = \sum_{X,Y \in S} \sum_{i=1}^{|Y|} \text{log} P(Y_i|Y_{<i},X)
\label{equ: learning obj}
\end{equation}
where $P$ indicates the probability distribution modelled by GPT-2 or T5, $|Y|$ is the length of the label sentence and $Y_i$ is the $i$-th token of $Y$.




\subsection{Metric}


Following~\citet{shu-etal-2021-logic}, we use BLEC to measure the logical consistency between the logical forms and the generated texts. Given a test dataset $\mathbb{D}$, BLEC is defined as:
\begin{equation}\label{vanilla BLEC}
    \mathrm{BLEC} = \frac{\sum_{s \in \mathbb{D}} \mathbb{I}_s(\text{oper , num}) }{|\mathbb{D}|}
\end{equation}
where function $\mathbb{I}_s(\text{oper}, \text{num})$ checks whether the \textbf{oper}ators (oper) or \textbf{num}bers (num) of the logical form can be exactly found in the sentence generated by the model. $|\mathbb{D}|$ is the size of $\mathbb{D}$. However, this implementation ignores checking the table headers of the logical form, only focusing on the accuracy of operators and numbers.
In order to take the table headers into consideration, we define an improved metric named BLEC* shown as follows:
\begin{equation}\label{BLEC*}
    \mathrm{BLEC^*} = \frac{\sum_{s \in \mathbb{D}} \mathbb{I}_s(\text{oper, num, header})}{|\mathbb{D}|}
\end{equation}

\subsection{Results}


GPT-2 and T5 based Logic2Text models are trained on the original training dataset, then they are evaluated on the standard test dataset (L2T) and LCD, respectively.

\begin{table}[t]
    \centering
    \begin{tabular}{p{1.1cm}p{0.6cm}p{0.6cm}p{0.6cm}p{0.6cm}p{0.6cm}p{0.6cm}}
    \toprule[1.5pt]
     \multirow{2}*{\textbf{Models}} & \multicolumn{3}{c}{\textbf{BLEC*}} & \multicolumn{3}{c}{\textbf{BLEC}}\\
     & \textbf{L2T} & \textbf{LCD} & \textbf{Dec} & \textbf{L2T}  & \textbf{LCD} & \textbf{Dec}\\
    \toprule[1.5pt]
     GPT-2 & 61.17 & 28.18 & 54\% & 83.52 & 58.96 & 29\%\\
    T5 & 71.61 & 41.78 & 42\% & 88.00 & 70.58 & 20\%\\
     \bottomrule[1.5pt]
\end{tabular}
    \caption{Baseline performance on test set of Logic2Text (L2T) and Logical Counterfactual Data (LCD). Dec denotes the relative decrease on LCD compared with L2T.}
    \label{tab: cusDataBaseline}
\end{table}

As shown in Table~\ref{tab: cusDataBaseline}, compared with the performance on L2T, there is a serious decline for both T5 and GPT-2 on LCD with respect to BLEC* and BLEC. Specifically, for T5, the BLEC* score decreases from 71.61 to 41.78, giving a relative decrease of 42\%. For GPT-2, the model achieves a BLEC* score of 61.17 on L2T, while only obtaining a BLEC* score of 28.18 on LCD, giving a relative decrease over 50\%.
Since BLEC only checks the accuracy on operators and numbers, the decline of BLEC is relatively small.
As seen, T5 performs slightly better than GPT-2 on LCD.
The reason we suppose is that since GPT-2 is an autoregressive language model, when used as an encoder it can only see partial logical forms.

Results on LCD verify our hypothesis that the models trained on the biased dataset learn to exploit the spurious correlations to perform logical reasoning, which are not robust when encountering counterfactual samples.

\begin{figure}[t]
    \centering
    \includegraphics[width=\linewidth]{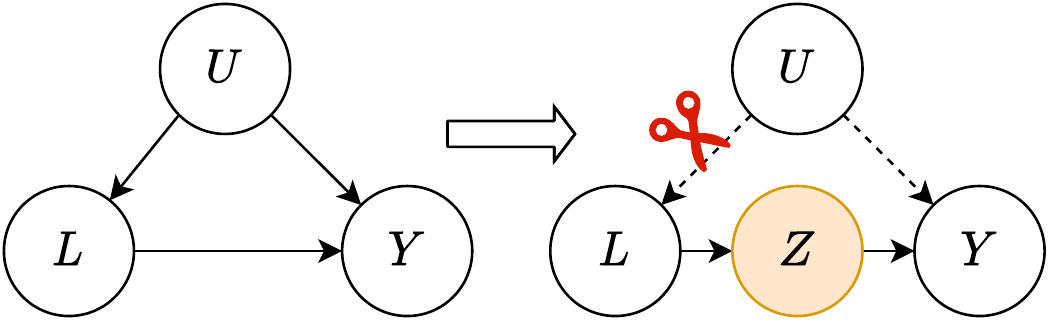}
    \caption{Causal graph for the task of Logic2Text.
    Left: The original method. Right: Our proposed modifications.}
    \label{causal graph}
\end{figure}

\section{Causal Analysis}
To deal with this problem, we utilize \textit{Causal Inference}~\cite{2016Causal} to analyze this bias. The left graph of Figure \ref{causal graph} illustrates the overall training process of conventional Logic2Text models from a causal perspective, where vertex $L$ and $Y$ denote the linearized logical form and the label sentence, respectively. Vertex $U$ represents an unobserved confounder in Logic2Text~\footnote{A \textit{Confounder} is a variable that influences both the treatment and the outcome, causing a spurious association. Confounding is a causal concept, and as such, cannot be described in terms of correlations or associations \cite{VanderWeele_2013}.}, which we suppose is the preference of the annotators to describe a table. There are also three edges in the left causal graph: $U \rightarrow L$, $U \rightarrow Y$ and $L \rightarrow Y$.
Edge $U \rightarrow L$ and $U \rightarrow Y$ denote the confounder's effects on $L$ and $Y$, respectively. $L \rightarrow Y$ represents that the label sentence is dependent on the logical form $L$. This link is the objective that the model should learn.

Concretely, in the Logic2Text task, the confounder $U$ can be interpreted as the preference of the annotators to describe a table. For example, given a table recording sports events, the annotators prefer to describe the game with the largest crowd attendance rather than the most recent game. 
As a consequence, this unobserved confounder has direct effects on the generation of the logical form ($U \rightarrow L$), and the generation of the label sentence ($U \rightarrow Y$). These effects finally build a backdoor path from the logical form to the label sentence ($L \leftarrow U \rightarrow Y$). The backdoor path induces the model to learn the shortcut between the logical form and the label sentence, rather than the process of reasoning sentence from the real structure of logical form ($L \rightarrow Y$). For example, when a model trained on a dataset in which a large number of logical forms containing ``$\text{ argmax \{ all}\_\text{rows ; attendance \}}$'',  the model can make correct predictions ``the highest attendance''. However, when testing on the logical form ``$\text{ argmax \{ all}\_\text{rows ; date \}}$'', the model leverages this shortcut and still employs the “argmax” logical operator to describe “attendance”, which is inconsistent with the logical form. 

Formally, given a confounder $U$, there exists a logical operator $o_i \in O$ and a table header $h_j \in H$ such that $p(o_i,h_j| U)$ is much higher than $p(o_i, h_k| U)$ for other $h_k \in H$, where $O, H$ represent the set of logical operators and table headers in the dataset respectively. $p(o,h)$ denotes the probability that the header $h$ should reason with operator $o$. In this case, the models actually learns the probability $P(Y|U, L, T)$, rather than $P(Y|L,T)$.

\section{Methodology}
Based on the above analysis, we make two modifications on the causal graph as shown on the right part of Figure \ref{causal graph}. Firstly, we propose structure-aware logical form encoder to build $L \rightarrow Z \rightarrow Y$. Secondly, we remove the backdoor path $L \leftarrow U \rightarrow Y$ by training the model on automatically generated counterfactual samples.

\subsection{Structure-Aware Logical Form Encoder by Building $L \rightarrow Z \rightarrow Y$}
\label{sec:incoporate_structrue}
In order to overcome the drawbacks of the linearized logical form inputs in previous works, a vertex $Z$ is added to the causal graph, which represents the structure-aware feature of the logical form. Then, we replace the edge $L \rightarrow Y$ with the path $L \rightarrow Z \rightarrow Y$, as shown in the right part of Figure~\ref{causal graph}. In the implementation, this modification is achieved by using an attention mask matrix to constrain each token in the logical form to only interact with the tokens it should reason with.

Specifically, let $L$ denote a linearized logical form $L=\{w_1, w_2,...w_n\}$ where $w_i \in V \cup O$. $V$ and $O$ are the vocabulary and the set of logical operators, respectively. Let $\mathbf{M}$ denote the attention mask matrix. $\mathbf{M}_{i,j} = 0$ indicates that the attention value from word $w_i$ to $w_j$ is masked, otherwise the opposite. Take the logical form ``$\text{hop} \{ \text{ argmax \{ all}\_\text{rows ; attendance \}; date }\}$'' as an example. The attention mask vector for the logical operator $\text{``argmax"}$ is $(0, 0, 1, 1, 1, 1, 1, 1, 0, 0, 0)$. With this constraint, the model will not be able to learn the association between $\text{``argmax''}$ and $\text{``date''}$. 

To decide the values in the attention mask matrix $\mathbf{M}$, we need convert the linearized logical form $L$ into a logical graph $L_{g}$.
To be more specific, for each token in the linearized logical form, there is a corresponding node in $L_g$. If a logical operator $w_i$ and a table header $w_j$ satisfy the pattern ``$w_i$ \{ $A$ ; $w_j$ ; $B$ \}'' in the linearized logical form, where $A$ and $B$ denote valid logical subclauses, then $w_i$ and $w_j$ in $L_g$ are connected.

Formally, let $\mathbf{M}_{i,j} = \text{edge}(w_i,w_j)$, where $\text{edge}$ is a binary function which indicates whether there is an edge between word $w_i$ and $w_j$ in $L_g$. Based on $\mathbf{M}$, the attention matrix in each transformer layer of the pre-trained models is calculated as:
\begin{equation}
    \hat{\mathbf{A}} = \text{softmax}(\mathbf{M} \odot \mathbf{A})
\end{equation}
where $\mathbf{A}$ denotes the original attention values, $\odot$ represents the element-wise product. Based on $\hat{\mathbf{A}}$, the standard self-attention transformer can be performed to calculate the representations for each token.

\begin{figure}[t]
    \centering
    \includegraphics[width=\linewidth]{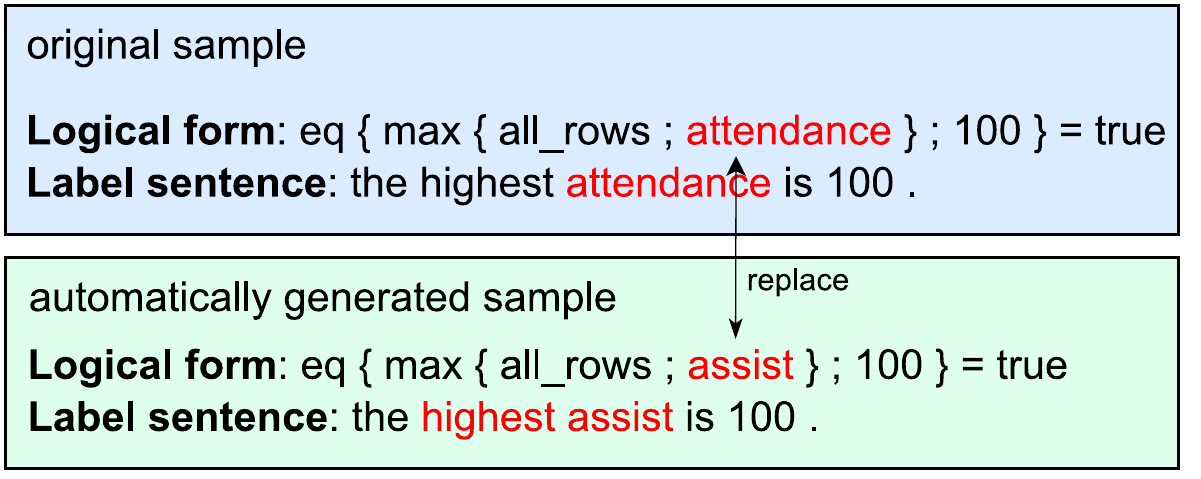}
    \caption{Automatically generated counterfactual sample for training.}
    \label{fig: autogen_CF}
\end{figure}

\subsection{Counterfactual Data Synthesizing to Remove the Backdoor Path $L \leftarrow U \rightarrow Y$}\label{backdoor-adj}
\label{sec:cut_backdoor}
In order to constrain the model to only rely on the direct path $L \rightarrow Y$ to generate sentences, the second modification is to remove the backdoor path $L \leftarrow U \rightarrow Y$ in the left causal graph of Figure~\ref{causal graph}. The backdoor path induces the model to learn the shortcuts between the logical form and the label sentence. This modification is implemented by training the model on automatically generated counterfactual samples. 



Specifically, given a selected table header in the logical form, we propose to replace it with another table header, which is randomly selected from the set of all the table headers in the training dataset. Accordingly, the exactly same tokens in the label sentence are also replaced with the selected table header. This replacement constructs counterfactual samples containing some rarely co-occurred headers and logic operators, which violates the preference of the annotators, thus removing the edges from the unobserved confounder $U$ to $L$ and $Y$. Take Figure \ref{fig: autogen_CF} as an example. Given a linearized logical form, and the corresponding label sentence, we locate the table header ``attendance'' and replace it with another randomly selected table header ``assist''. We also replace ``attendance'' with ``assist'' in the label sentence. We filter out samples whose logical forms do not contain table headers that can exactly match any tokens in the label sentence. Due to the space limitation, three strategies we used to select the table headers to be substituted are listed in the Appendix \ref{appendix: cfdata gene method}.


Based on the automatically generated counterfactual data $\tilde{S}$, the model is trained on the mixup of the counterfactual data $\tilde{S}$ and the original training dataset $S$ as follows:
\begin{equation}
    J = \sum_{X,Y \in \tilde{S} \cup S} \sum_{i=1}^{|Y|} \text{log} P(Y_i|Y_{<i},X)
\end{equation}
It is worth noting that the label sentences of the automatically generated counterfactual dataset may not be natural sentences, since the randomly selected table headers may not fit the contexts of the original label sentences. As a result, adding more automatically generated counterfactual dataset would improve the logical consistency of the generated texts, but the fluency of the texts may be hurt. The trade-off is decided by the ratio $r$ between the size of $\tilde{S}$ and the size of $S$. $r$ can be calculated as $|\tilde{S}| / |S|$. An experiment of exploring the effect of $r$ can be found at Subsection \ref{rate-exp}.




\section{Experiments}\label{sec: experiments}

In this section we conduct experiments on both the biased Logic2Text dataset and the counterfactual dataset, LCD. We compare our method with other SOTA models, and discuss the experimental results.

\paragraph{Baselines}
We use 1) the Pointer Generator Network \cite{see-etal-2017-get}, 2) SNOWBALL \cite{shu-etal-2021-logic}, 3) GPT-2 \cite{radford2019language} and 4) T5 \cite{T5} as baselines. The details about the baselines are shown in Appendix \ref{appendix: baselines}.

\begin{table*}[t]
    \centering
    \begin{tabular}{cccccccccc}
    \toprule[1.5pt]
        \multirow{2}*{\textbf{Method}} & \multicolumn{3}{c}{\textbf{BLEC*}} & \multicolumn{3}{c}{\textbf{BLEC}} & \multicolumn{3}{c}{\textbf{Human Evaluation}} \\
        & \textbf{L2T $\uparrow$} & \textbf{LCD $\uparrow$} & \textbf{Dec $\downarrow$} & \textbf{L2T $\uparrow$}  & \textbf{LCD $\uparrow$}  & \textbf{Dec $\downarrow$}  & \textbf{L2T $\uparrow$}  & \textbf{LCD $\uparrow$}  & \textbf{Dec $\downarrow$} \\
        \midrule[1.5pt]
        Pointer Network & 40.11 & 11.62 & 71\% & 70.42 & 33.13 & 53\% & 51 & 13 & 75\%  \\
        SNOWBALL & 72.89 & 51.92 & 29\% & 85.35 & \textbf{82.94} & \textbf{3}\% & 77 & 52 & 32\% \\
        \hline
        GPT-2 & 61.17 & 28.18 & 54\% & 83.52 & 58.96 & 29\% & 71 & 31 & 56\% \\
        T5 & 71.61 & 41.78 & 42\% & 88.00 & 70.58 & 20\% & 83 & 41 & 51\%\\
        \hline
        GPT-2 + Ours & 71.70 & 48.83 & 32\% & 87.72 & 66.74 & 24\% & 79 & 43 & 46\%\\
        T5 + Ours & \textbf{83.42} &	\textbf{59.58} & \textbf{28}\% & \textbf{89.93} &	72.68 & 19\% & \textbf{84} & \textbf{71} & \textbf{15\%} \\
        \bottomrule[1.5pt]
    \end{tabular}
    \caption{Results of different methods on L2T and LCD with respect to BLEC, BLEC* and human evaluation. Dec denotes the relative decrease when estimated on LCD compared with on L2T, which is calculated as $\frac{L2T-LCD}{L2T}$. The up arrow $\uparrow$ means that the metric is as higher as better, the down arrow $\downarrow$ is the opposite.}
    \label{tab: main result}
\end{table*}

\paragraph{Implementation Details}
In the experiments, the learning rate is set to 0.0003. We use beam search for decoding and the beam size is set to 2. The maximum length of the output sentence is set to 180. The batch size is set to 10 for inference and 2 for training. For the T5 backbone, we initialized the parameters from CodeT5 \cite{wang-etal-2021-codet5}, which
is pre-trained on programming languages. Linearized logical forms contain similar structures, with operators and table headers corresponding to functions and parameters, respectively.
For SNOWBALL, we follow the settings in \cite{shu-etal-2021-logic} but reduce the batch size and beam size to 4 due to the limitation of GPU memory. We train all the models on one GeForce GTX 1080 Ti.

\paragraph{Manual Evaluation}
In addition to the automatic evaluation, we manually check the logical consistency by comparing the output sentences with the logical forms. Specifically, we randomly select 100 samples from L2T and LCD, then we calculate the percentage of the samples that the output sentence is logically consistent with the logical form. 

\paragraph{Relative Decrease}

The LCD and L2T have comparable data sizes, with 809 and 1092 samples, respectively. Here We use relative \textbf{Dec}rease to quantify the degree of decline in model performance when test data is transferred from L2T to LCD, which is calculated as $(L2T - LCD) / L2T$. $L2T$ denotes the model performance on standard Logic2Text test set, $LCD$ denotes the performance on LCD.

\subsection{Main Results}
\label{sec:main result}

As shown in Table \ref{tab: main result}, augmenting GPT-2 and T5 with our method obtains a large gain on both L2T and LCD with respect to BLEC* and BLEC. 

Specifically, for T5, the BLEC* score increases to 83.42 on L2T and 59.58 on LCD, which outperforms all the baselines and achieves the new state-of-the-art results.

For GPT-2 with our method, the BLEC* increases by 20.65 on LCD and 10.53 on L2T. The relative decrease is reduced from 54\% to 32\%. Both the results of GPT-2 and T5 demonstrate that our modifications to the causal graph are effective. For BLEC, T5 with our method obtains the highest BLEC score of 89.93 on L2T, which shows a slight improvement compared with vanilla T5 (88.0). For GPT-2 with our method, the BLEC score increases by 4.2 on L2T and 7.78 on LCD. The improvement of BLEC is relatively limited compared with BLEC*. The reason we suppose is that BLEC only checks the accuracy of operators and numbers, which cannot reveal the errors of table headers. This problem is also demonstrated by the case analyses of SNOWBALL. Although SNOWBALL gives a low decrease, when looking into some cases generated by SNOWBALL, we find that SNOWBALL fails to choose the correct table headers.

A similar conclusion can be drawn from human evaluation. Our T5-based method achieves the highest accuracy on L2T (84) and LCD (71), giving the lowest relative decrease (15\%). Compared with vanilla T5, the manually checked logical accuracy of our method obtains a 30-point improvement on LCD, which further verifies the effectiveness of our method.

\begin{table}[t]
    \centering
    \footnotesize
    \setlength\tabcolsep{3pt}
    \begin{tabular}{ccccc}
    \toprule[1.5pt]
    \multirow{2}*{$r$} & \multicolumn{2}{c}{\textbf{L2T}}
    & \multicolumn{2}{c}{\textbf{LCD}}\\
	& \textbf{BLEC*} /	\textbf{BLEC} &	\textbf{BLEU} &	\textbf{BLEC*} / \textbf{BLEC} & \textbf{BLEU} \\
	\toprule[1.5pt]
    0&	71.61 /	88& 	30.41&	41.78 /	70.58&	23.69	\\
    1&	74.82 /	88.28&	30.35&	46.97 /	72.44&	21.85	\\
    5&	79.69 /	90.93&	28.64&	56.98 /	76.02&	22.79	\\
    10&	79.3 /	89.65&	26.89&	57.35 /	72.44&	22.05	\\
    20&	82.88 /	90.84&	28.02&	59.46 /	73.3&	22.29	\\
    $\infty$&	83.42 /	89.93&	27.19&	59.58 /	72.68&	23.68\\
    \toprule[1.5pt]
    \end{tabular}
    \caption{The effect of the proportion of synthetic training data on the quality of the generated text. $r = |\tilde{S}| / |S| $, where $|\tilde{S}|$ denotes the size of counterfactual data, $|S|$ denotes the size of the original training data. $r=\infty$ indicates that all of the training data are synthetic counterfactual samples.}
    \label{rate}
\end{table}

\subsection{Effect of the Size of Synthetic Data}\label{rate-exp}

\Chengyuan{the extra experiment}
As indicated in Section \ref{backdoor-adj}, adding more automatically generated counterfactual examples would improve the logical consistency of the generated texts, but the fluency of the texts may be hurt. The trade-off is decided by the ratio between the size of counterfactual data $\tilde{S}$ and the size of original training data $S$. We conduct development experiments to quantitatively explore the effect of the trade-off on the logical consistency and text fluency. We use BLEC* and BLEC to evaluate the logic consistency, and BLEU-4\footnote{Standard script NIST mteval-v13a.pl}\cite{papineni2002bleu}(noted as BLEU) to evaluate the text fluency.

As shown in Table \ref{rate}, with the increase of $r=|\tilde{S}| / |S|$, the logical correctness increases and the text fluency decreases, which verifies our hypothesis in Section~\ref{backdoor-adj}. We suggest that the future researchers fine-tune the hyperparameter $r$ to obtain logically correct and semantically fluent generated texts.


\subsection{Effect of the Complexity of Logical Form}\label{sec:complexity of lf}

We explore how the model performance is affected by the complexity of the logical form. The effects of the complexity are demonstrated from two aspects: 1) the maximum depth of the logical form in a tree form, and 2) the number of nodes in the logical form. The logical form with more nodes or deeper depth is regarded as more complex. 

\begin{figure}[t]
    \centering
    \subfigure[Mispredicted tokens rate with respect to the number of nodes in the logical form.]{
        \includegraphics[width=\linewidth]{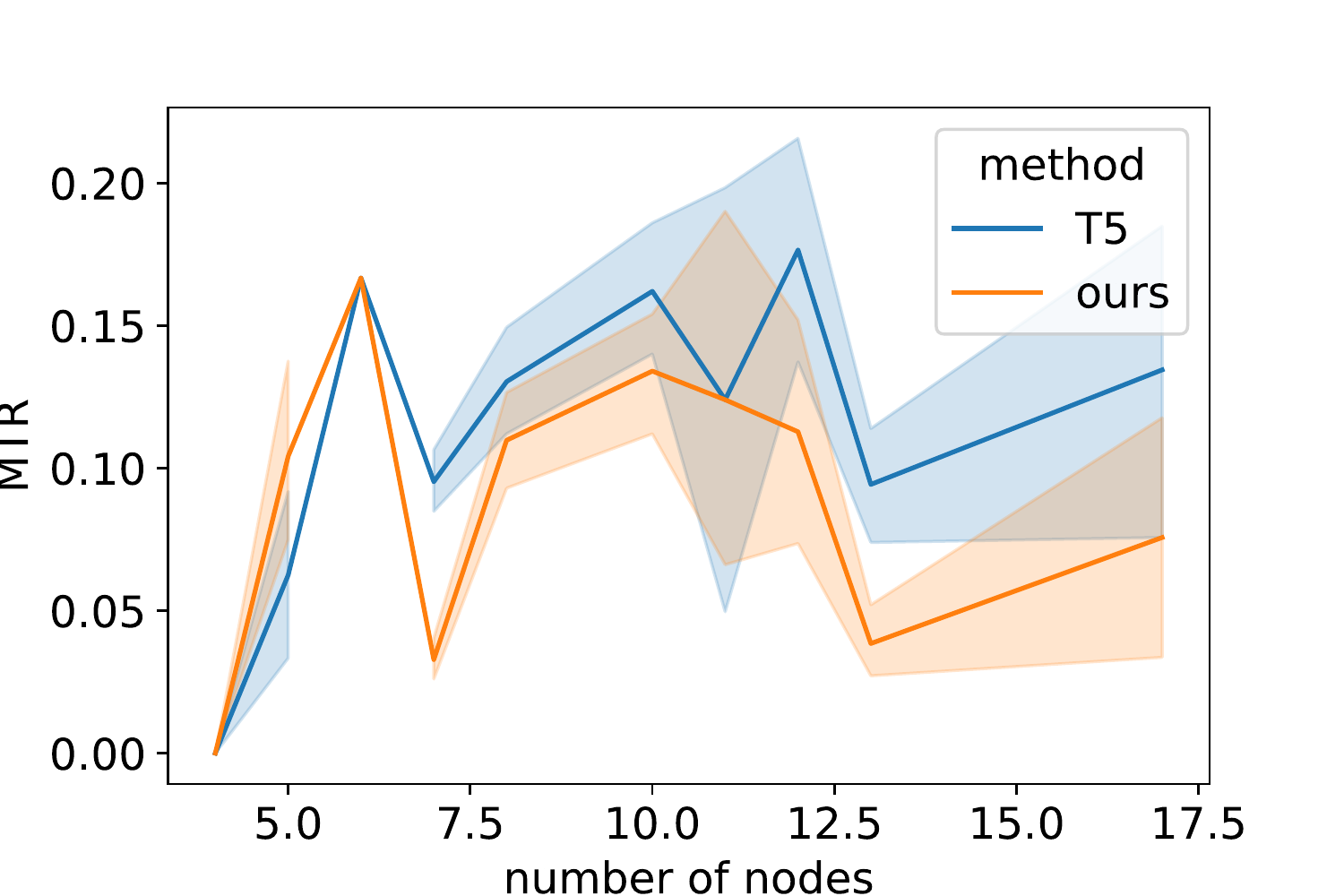}
        \label{fig:complexity:nodes}
    }
    \subfigure[Mispredicted tokens rate with respect to the depth of the logical form.]{
        \includegraphics[width=\linewidth]{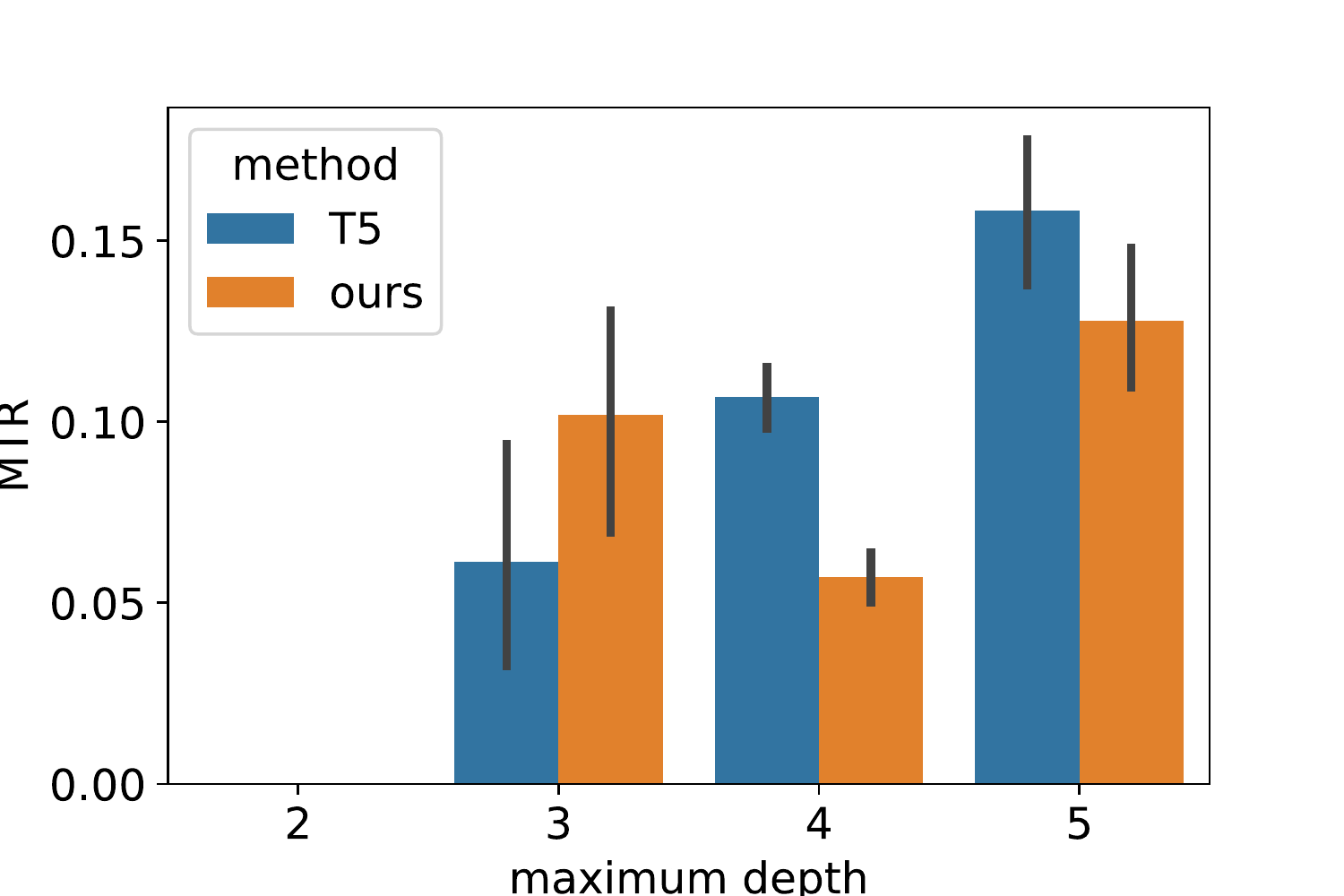}
        \label{fig:complexity:depth}
    }
    \caption{Mispredicted token percentage under various complexity.}
    \label{fig: complexity}
\end{figure}

We take T5 as the backbone and report the mispredicted token rate (MTR) as the metric. Mispredicted tokens are the tokens that occur in the logical form but not generated in the text. We calculate the ratio of the number of the mispredicted tokens to the length of the logical form. We plot the MTR results with respect to different numbers of nodes and depths in Figure~\ref{fig: complexity}.

As shown in Figure~\ref{fig:complexity:nodes}, as the number of nodes increases, more tokens are mispredicted by the vanilla T5 model. In contrast, our method helps T5 to maintain a relatively low error rate as the number of nodes increases. A similar conclusion can be drawn from Figure~\ref{fig:complexity:depth}. Our approach makes the model stable as the logical depth grows. Besides, we observe that for logical forms with low depths, our method obtains a slight higher MTR compared with the baseline. The reason we suppose is that simple logical forms are rarely affected by the unobserved confounder.

\begin{table*}[t!]
    \centering
    \begin{tabular}{lccccccc}
    \toprule[1.5pt]
        \multirow{2}*{\textbf{Method}} & \multirow{2}*{\textbf{
        \begin{tabular}[c]{@{}l@{}}
        Backbone
        \end{tabular}
        }} & \multicolumn{3}{c}{\textbf{BLEC*}} & \multicolumn{3}{c}{\textbf{BLEC}} \\
         & & \textbf{L2T}$\uparrow$ & \textbf{LCD}$\uparrow$ & \textbf{Dec}$\downarrow$ & \textbf{L2T}$\uparrow$ & \textbf{LCD}$\uparrow$ & \textbf{Dec}$\downarrow$ \\
         \midrule[1.5pt]
        w/o both & \multirow{4}*{T5} & 71.61 & 41.78 & 42\% & 88.00 & 70.58 & 20\% \\
        w/o CF & & 65.93 & 39.80 & 40\% & 87.82 & 66.13 & 25\% \\
        w/o AM  & & 69.87 & 47.84 & 32\% & 87.91 & \textbf{74.16} & \textbf{16\%} \\
        FULL & & \textbf{83.42} & \textbf{59.58} & \textbf{28\%} & \textbf{89.93} & 72.68 & 19\%\\
        \hline
        w/o both & \multirow{4}*{GPT-2} & 61.17 & 28.18 & 54\% & 83.52 & 58.96 & 29\% \\
        w/o CF & & 61.72 & 34.48 & 44\% & 85.16 & 66.25 & \textbf{22\%} \\
        w/o AM & & 69.60 & 42.15 & 39\% & 84.98 & 59.09 & 30\% \\
        FULL &  & \textbf{71.70} & \textbf{48.83} & \textbf{32\%}& \textbf{87.72} & \textbf{66.74} & 24\% \\
        \bottomrule[1.5pt]
    \end{tabular}
    \caption{Results of ablation study. We explore the effect of the attention mask (AM) and training on counterfactual data (CF).}
    \label{tab:ablation}
\end{table*}

\begin{table}
    \centering
    \begin{tabular}{ccccc}
        \toprule[1.5pt]
        \multirow{2}*{\textbf{Generation}} & \multicolumn{2}{c}{\textbf{BLEC*}} & \multicolumn{2}{c}{\textbf{BLEC}} \\
         & \textbf{L2T} & \textbf{LCD} & \textbf{L2T} & \textbf{LCD} \\
        \toprule[1.5pt]
        - & 65.93 & 39.80 & 87.82 & 66.13\\
        Random & 75.00 & 47.34 & 88.10 & 64.77 \\
        Disturb & 79.67 & 59.20 & 88.19 & 72.44 \\
        Mix & \textbf{83.42} & \textbf{59.58} & \textbf{89.93} & \textbf{72.68}\\
        \bottomrule[1.5pt]
    \end{tabular}
    \caption{Results of different methods of generation of counterfactual data vary on L2T and LCD, where ``-'' denotes training on standard Logic2Text.}
    \label{tab: generation method}
\end{table}

\subsection{Ablation Study}\label{sec:ablation study}
We conduct ablation experiments on both T5 and GPT-2 by removing the attention mask (denoted as AM) and automatically generated counterfactual training data (denoted as CF). The results are shown in Table \ref{tab:ablation}.

When removing AM, for GPT-2, the BLEC* score decreases by 2.1 and 6.68 on L2T and LCD respectively. The BLEC score decreases by 2.98 on L2T and 7.65 on LCD.
For T5, removing AM leads to 13.55 and 11.74 decreases on L2T and LCD, respectively. These observations demonstrate the effect of AM.

When removing the automatically generated counterfactual data for training, the performances for both GPT-2 and T5 are significantly decreased.
Specifically, for GPT-2, the BLEC* decreases by 9.98 on L2T and 14.45 on LCD.
For T5, the BLEC* decreases by 17.49 on L2T and 19.78 on LCD. The decline increases from 28\% to 40\%.
There is much worse decrease on LCD than L2T since the data distribution of LCD is different from L2T which shares the same data distribution with the training data.

\subsection{Generation of Counterfactual Data}\label{sec:exp counterfactual data}

We conduct experiments to investigate the effects of different strategies to generate counterfactual data. Specifically, we try to replace the table header tokens in the logical form, with 1) a random string (denoted as Random), 2) a randomly selected table header (denoted as Disturb), 3) and mix the above two methods up (denoted as Mix). We list the details of the methods in Appendix \ref{appendix: cfdata gene method}. The results are shown in Table \ref{tab: generation method}. 

We observe that Disturb greatly boosts the performance more than Random on both L2T and LCD, which proves that meaningful and in-domain table headers generate more effective counterfactual samples. Besides, the Mix strategy further give a slightly improvement.

\section{Case Study}\label{sec: case study}

We select a counterfactual sample from LCD to demonstrate why our method could perform better than previous works using the attention score distributions when decoding.
The linearized logical form of the sample is ``eq\{ hop \{argmax \{all\_rows ; score \};attendance \};5032\}'', and the label sentence is ``the game with the highest score had 5032 spectators''.

\begin{figure}
    \centering
    \includegraphics[width=\linewidth]{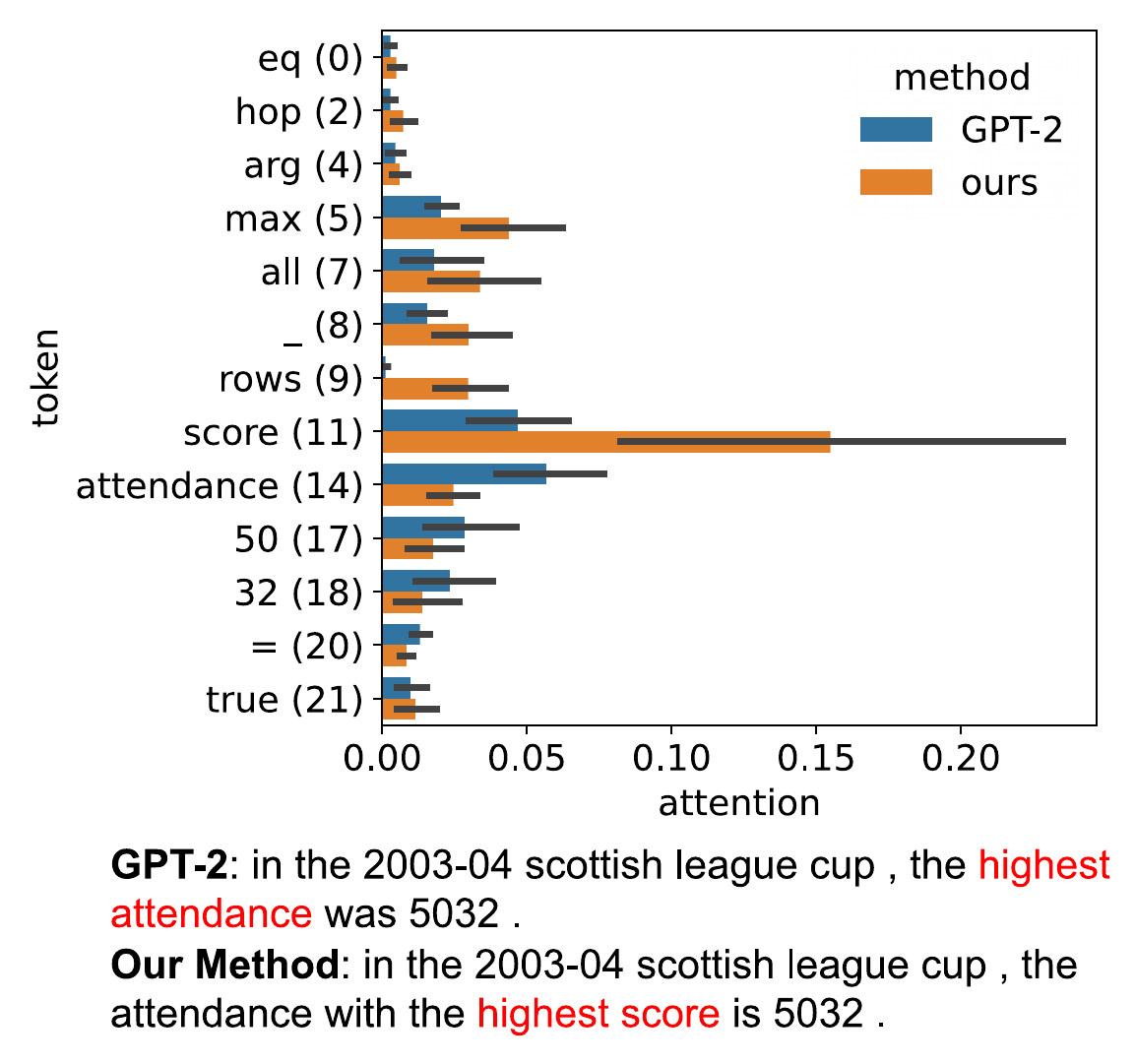}
    \caption{Attention values during decoding. The baseline pays more attention to ``attendance'' as we expected, which verifies our hypothesis.}
    \label{fig: case study}
\end{figure}

Specifically, after decoding the ``highest'' or ``largest'' token corresponding to the ``argmax" logical operator in the logical form, we plot the attention values of the last transformer layer for each token of the logical form to explore which table header the model would choose, as shown in Figure~\ref{fig: case study}. As seen, for vanilla GPT-2, the attention score of the token ``attendance'' is higher than on ``score'', thus producing ``highest attendance''. For our method, the attention score of the token ``score'' achieves the largest, which is the corresponding table header the operator ``argmax'' should select. 
\section{Related Work}\label{sec: related work}

\subsection{Text Generation from Tables}

Table-to-text is a popular area in recent years~\cite{wiseman-etal-2018-learning, DBLP:journals/corr/abs-1806-01353, liang-etal-2009-learning, chen2021confounded}. As previous methods generate superfacial and uncontrollable logic, \citet{chen-etal-2020-logic2text} introduced Logic2Text as a controllable and fidelity text generation task conditioned on a logical form.
Since then, many works on Logic2Text have been proposed.
In order to unify the studies of structural knowledge grounding, \citet{xie2022unifiedskg} proposed the UNIFIEDSKG framework and unified 21 structural knowledge grounding tasks into a text-to-text format, including Logic2Text.
\citet{logen} proposed a unified framework for logical knowledge-conditioned text generation in few shot setting.  To solve the data scarcity problem of Logic2Text, \citet{shu-etal-2021-logic} iteratively augmented the original dataset with a generator and proposed an evaluator for high-fidelity text generation.

However, they all ignored the spurious correlation in logical forms, which is investigated in our work.

\subsection{Causal Inference For NLP}

Causal Inference \cite{2016Causal, kuang2020causal} is a powerful statistical modeling tool for explanatory analysis.
In NLP, many methods have been proposed based on the causal inference theory~\cite{zhang2021causerec, DBLP:conf/cvpr/0016YXZPZ20,zhang-etal-2021-de,hu2021causal}.
\citet{yang-etal-2021-exploring} and \citet{DBLP:conf/aaai/WangC21} exploit causal inference to reduce the bias from the context for text classification tasks.
For named entity recognition, \citet{zeng-etal-2020-counterfactual} replaced the entities in sentences with counterfactual tokens to remove spurious correlation between the context and the entity token. 
\citet{wang2021robustness} generated counterfactual samples by replacing causal terms with their antonyms in sentiment classification. \citet{wu-etal-2020-de} proposed to use a counterfacutal decoder to generate unbiased court's view.

Our work proposes to improve the robustness of Logic2Text models with causality.

\section{Conclusion}
We investigate the robustness of current methods for Logic2Text via a set of manually constructed counterfactual samples.
A significant decline on the counterfactual dataset verifies the existence of bias in the training dataset.
Then we leverage causal inference to analyze the bias, based on which, two approaches are proposed to reduce the spurious correlations.
Automatic and manual experimental results on both Logic2Text and the counterfactual data demonstrate that our method is effective to alleviate the spurious correlations.

\section*{Limitations}



Although our method has achieved high logical consistency, we find that for some unseen headers, the model cannot understand them and generate some logically correct but not fluent sentences, which is related to the method of generation of counterfactual samples. Due to the limited number of high-quality logical forms, future work may continue to explore more advanced counterfactual data generation methods considering the context.

Besides, our structure-aware logical form encoder works based on the attention mechanism, so it can't be applied to models without attention. Fortunately, the current attention-based models are widely used not only because of their better performance but also because of their high interpretability.

\section*{Acknowledgment}
We would like to thank anonymous reviewers for their comments and suggestions. This work is supported in part by National Natural Science Foundation of China (NO. 62037001), the Key R\&D Projects of the Ministry of Science and Technology (NO. 2020 YFC 0832500), the Zhejiang Province Science and Technology Project (NO. 2022C01044), the Starry Night Science Fund of Zhejiang University Shanghai Institute for Advanced Study (NO. SN-ZJU-SIAS-0010), and CAAI-Huawei MindSpore Open Fund (NO. CAAIXSJLJJ-2021-015A).

\bibliography{custom}
\bibliographystyle{acl_natbib}

\appendix






\begin{figure*}
    \centering
    \includegraphics[width=0.8\linewidth]{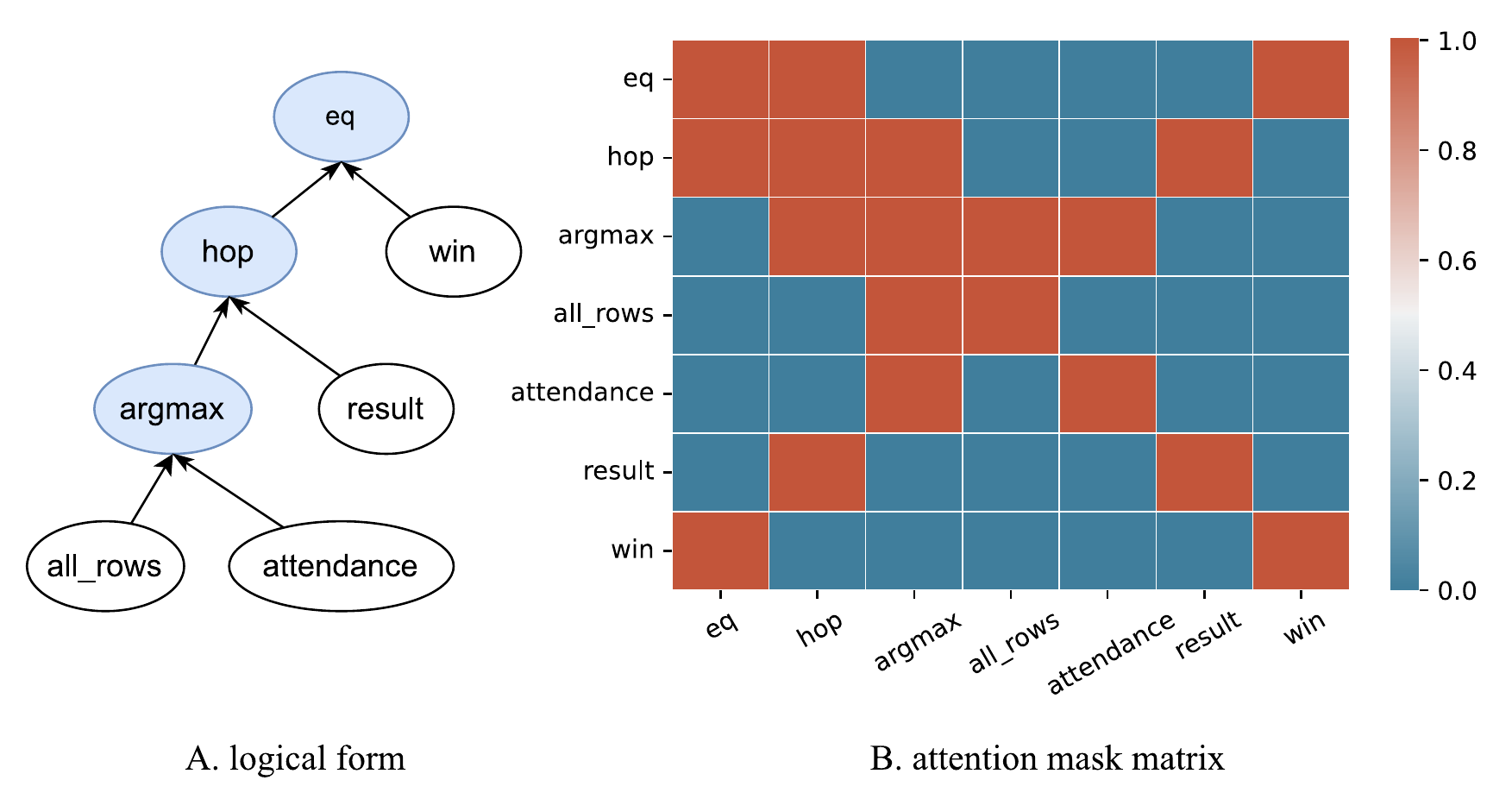}
    \caption{Sample of Attention Mask matrix. The attention of each token to others with no directly connected edges is masked.}
    \label{attnmaskFigure}
\end{figure*}

\section{Details of Attention Mask}\label{appendix: attn mask}

The attention value from token $w_i$ to token $w_j$ is masked if there is no direct edge connecting them on the logical form.
To clarify how the value of the Attention Mask is calculated, we use the left logical form in Figure \ref{attnmaskFigure} as an example. And the attention mask matrix for the tokenized logical form is shown on the right of Figure \ref{attnmaskFigure}.
For each token in the logical form, the parent node can be seen (such as $\mathbf{M}_{\text{hop}, \text{result}} = 1$). Besides, an operator token can also see the child nodes (such as $\mathbf{M}_{\text{win}, \text{eq}} = 1$).
Otherwise, the attention value is masked.

\section{Replacement Methods}\label{appendix: cfdata gene method}

We match the headers from each logical form to the tokens in the label and then replace the headers in a specific way if found.
Concretely, we propose the following three strategies for replacement.

\paragraph{Random Replacement}

Intuitively, when a layman tries to describe some domain-specific table, he simply replicates the obscure table headers (such as technical terms). So we train the model's ability to replicate the header from the logical form. We use completely random strings to replace the headers.

\paragraph{Header Disturb}

Another straightforward idea is to select another header token from a set of all table headers to replace the header token in the logical form. However, such a method ignores the attribute of the data type carried by the columns, thus it will produce unreasonable counterfactual samples. In order to solve this problem, we group all the headers by their data type, including three categories: strings, numbers, and time. A header in the logical form is only replaced by another header with the same data type.

\paragraph{Mixing Replacement}

We take turns performing the above two replacement strategies.

\section{Details of Baselines}\label{appendix: baselines}

\paragraph{Pointer Generator Network}
(PGN) \cite{see-etal-2017-get} can be employed to solve OOV problem. In addition to calculating the probability of each token in the existing vocabulary $P_{vocab}$, PGN also calculates $p_{gen}$ while decoding, where $p_{gen}$ denotes the probability to copy the tokens from the input sequence. The final distribution is calculated as:
\begin{equation}\label{pgn calculation}
    P(w) = p_{gen} P_{vocab}(w) + (1 - p_{gen})\sum_{i:w_i = w}a_t^i
\end{equation}

\paragraph{SNOWBALL}
To solve the constraint of data scarcity, \citet{shu-etal-2021-logic} proposed the SNOWBALL framework for high-fidelity text generation, which employed an iterative training procedure over a generator and an evaluator through data augmentation.

\paragraph{GPT-2}
\citet{radford2019language} proposed the left-to-right unidirectional generative model, with only the decoder of the transformer \cite{DBLP:conf/nips/VaswaniSPUJGKP17}.

\paragraph{T5}
\citet{T5} proposed the pre-trained model specifically for text-to-text generation tasks. We initialized the parameters from CodeT5~\cite{wang-etal-2021-codet5}, which is more suitable for formal-language-to-text.


\section{Effect On the Probability Of Copying Table Headers}
\label{sec: appendix improvements on l2t}

\begin{figure}[htb]
    \centering
    \includegraphics[width=\linewidth]{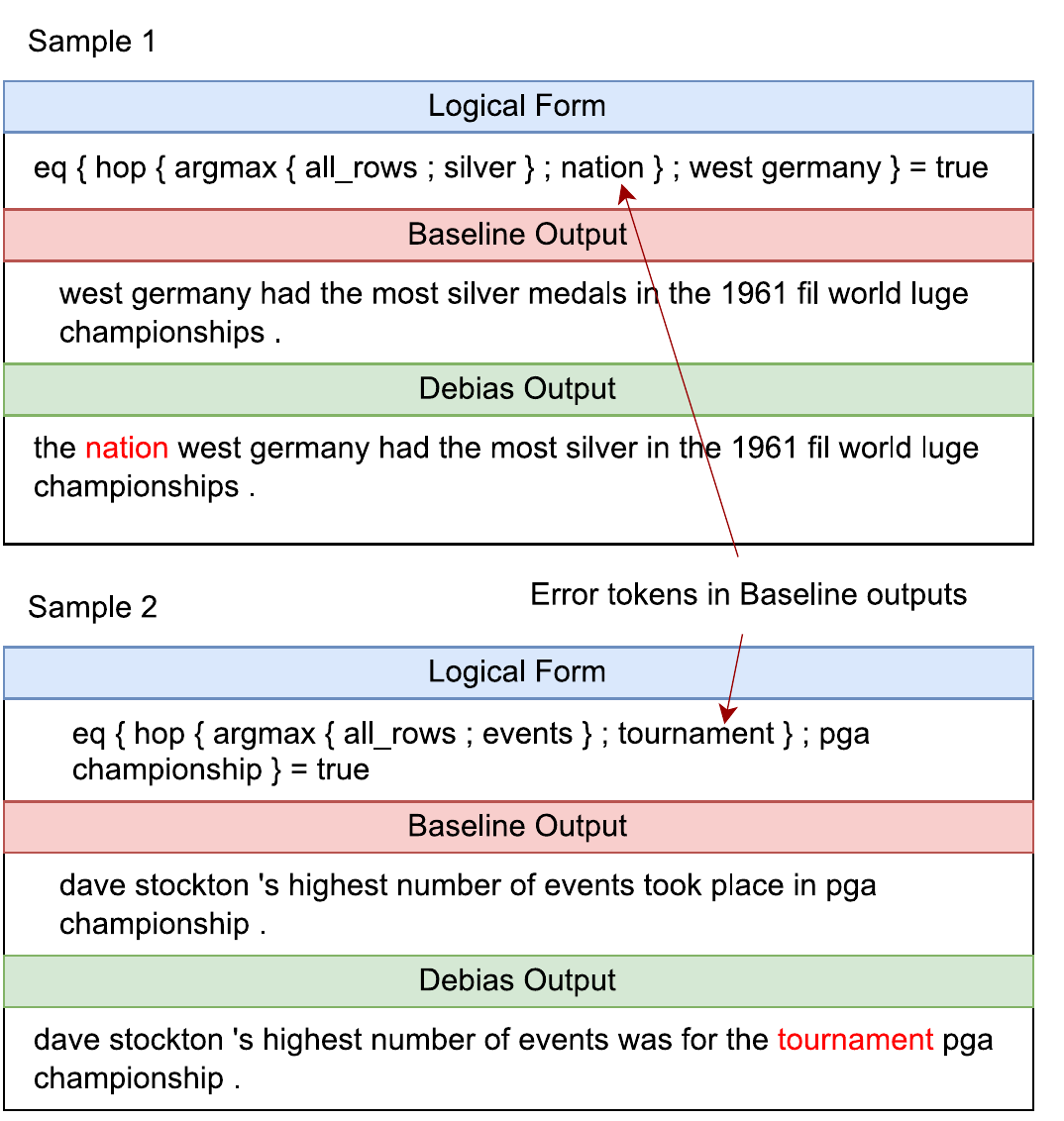}
    \caption{Our method contributes to the improvements on L2T.}
    \label{what improves l2t}
\end{figure}

Our method reduces the reliance on spurious correlations, making the model learn the relationship between key tokens. We find that our method increases the probability of copying headers from the logical forms, which affects the performance of the model on L2T as well.
We show two samples in Figure \ref{what improves l2t}.

In sample 1, the baseline just omits the header token ``nation'' because ``west germany'' itself is an instance of ``nation''. In sample 2, ``pga championship'' and ``tournaments'' have a similar relationship. But our method prefers to keep these headers if allowed.

\section{More Counterfactual Samples and Running Examples}

\Chengyuan{Add more detailed description and more running examples.}
We present 3 more counterfactual samples for reference. We list them in Table \ref{moresample1} to \ref{moresample3}. The concerned pairs of logical operators and table headers in original samples and counterfactual samples are highlighted in blue and red respectively. The counterfactual samples are modified from the original samples, where parts of the logical form are replaced with rarely co-occurred logic operators and table headers. Then the corresponding label sentences are re-edited.

\begin{table*}[]
    \centering
    \begin{tabular}{cm{6cm}m{6cm}}
    \hline
        & Original Sample & Counterfactual Sample \\
        \hline
        Logical Form &
        eq \{ hop \{ \textcolor{blue}{argmax \{ all\_rows ; televote \}} ; song \} ; dj , take me away \} = true
        & eq \{ hop \textcolor{red}{\{ argmin \{ all\_rows ; televote \}} ; song \} ; tazi vecher \} = true \\
        \hline
        Label & the song dj , take me away recevied the largest percentage of televotes . & 
        the song ' tazi vecher ' got the least televote in bulgaria in the eurovision song contest 2008 .\\
        \hline
    \end{tabular}
    \caption{Counterfactual Sample 1.}
    \label{moresample1}
\end{table*}
\begin{table*}[]
    \centering
    \begin{tabular}{cm{6cm}m{6cm}}
    \hline
        & Original Sample & Counterfactual Sample \\
        \hline
        Logical Form &
        eq \{ hop \{ \textcolor{blue}{nth\_argmin \{ all\_rows ; react ; 2 \}} ; athlete \} ; jaysuma saidy ndure \} = true
        & eq \{ hop \{ \textcolor{red}{nth\_argmax \{ all\_rows ; react ; 2 \}} ; athlete \} ; paul hession \} = true \\
        \hline
        Label & jaysuma saidy ndure had the second shortest react time among the 2008 summer olympics men 's 200 metres athletes . & 
        paul hession ' react time was the second longest in all athletes .\\
        \hline
    \end{tabular}
    \caption{Counterfactual Sample 2.}
    \label{moresample2}
\end{table*}
\begin{table*}[]
    \centering
    \begin{tabular}{cm{6cm}m{6cm}}
    \hline
        & Original Sample & Counterfactual Sample \\
        \hline
        Logical Form &
        \textcolor{blue}{most\_greater} \{ filter\_eq \{ all\_rows ; site ; memorial stadium minneapolis , mn \} ; \textcolor{blue}{attendance} ; 24999 \} = true
        & eq \{ max \{ \textcolor{red}{filter\_less} \{ all\_rows ; \textcolor{red}{attendance} ; 30000 \} ; date \} ; 11 / 24 / 1928 \} = true \\
        \hline
        Label & in the 1928 minnesota golden gophers football under clarence spears , most of the games at memorial stadium minneapolis , mn drew more than 24,999 people . & 
        the most recent game of minnesota golden gophers football under clarence spears which drew less than 30000 attendance was held on 11 / 24 / 1928 .\\
        \hline
    \end{tabular}
    \caption{Counterfactual Sample 3.}
    \label{moresample3}
\end{table*}

\begin{table*}[]
    \centering
    \begin{tabular}{cm{6cm}m{6cm}}
        \toprule[1.5pt]
        & Example 1 & Example 2 \\
        \toprule[1.5pt]
        \textbf{Logical Form} & eq \{ count \{ filter\_less \{ filter\_greater \{ all\_rows ; year ; 1975 \} ; points ; 1 \} \} ; 5 \} = true & eq \{ max \{ filter\_less \{ all\_rows ; area km square ; 10 \} ; population \} ; 5845 \} = true\\
        \hline
        \textbf{Label} & there were 5 times that hans - joachim stuck won less than 1 point in the after 1975 . & the largest population for the populated places in guam whose area is less than 10 km square is 5845 .\\
        \hline
        \textbf{Baseline} & in the 1975 season of hans - joachim stuck , among the years that he participated in , 5 of them had points less than 1 . & in guam , the highest population in area km square is 5845 .\\
        \hline
        \textbf{Ours} & for hans - joachim stuck , when the year is over 1975 , there were 5 times that there were less than 1 points . & the highest population in guam with area km square less than 10 is 5845 .\\
        \toprule[1.5pt]
    \end{tabular}
    \caption{2 running examples to prove the effectiveness of our method.}
    \label{more-running-exmaples}
\end{table*}

To prove the efficiency of our method, we give two more running examples in Table \ref{more-running-exmaples}. The results in the table illustrate that the Baseline model is prone to text inconsistent with logical forms when feeding counterfactual samples as input, while our approach yields more robust logical reasoning.

\end{document}